\documentclass[runningheads]{llncs}
\usepackage{graphicx}

\usepackage{tikz}
\usepackage{comment}
\usepackage{amsmath,amssymb} 
\usepackage{color}
\usepackage{orcidlink}

\usepackage[accsupp]{axessibility}  

\begin{document}
\pagestyle{headings}
\mainmatter
\def\ECCVSubNumber{224}  

\title{Deep Semantic Manipulation of Facial Videos} 

\titlerunning{Deep Semantic Manipulation of Facial Videos}
%
\author{Girish Kumar Solanki\inst{1}\orcidlink{0000-0002-4493-1726} \and
Anastasios Roussos\inst{2,1}\orcidlink{0000-0001-6015-3357}\\
\email{troussos@ics.forth.gr}}
\authorrunning{G.K. Solanki \& A. Roussos}
%
\institute{College of Engineering, Mathematics \& Physical Sciences, University of Exeter, UK \\
\and
Institute of Computer Science, Foundation for Research \& Technology - Hellas (FORTH), Greece}

\maketitle

\begin{abstract}
Editing and manipulating facial features in videos is an interesting and important field of research with a plethora of applications, ranging from movie post-production and visual effects to realistic avatars for video games and virtual assistants. Our method supports semantic video manipulation based on neural rendering and 3D-based facial expression modelling. We focus on interactive manipulation of the videos by altering and controlling the facial expressions, achieving promising photorealistic results. The proposed method is based on a disentangled representation and estimation of the 3D facial shape and activity, providing the user with intuitive and easy-to-use control of the facial expressions in the input video. We also introduce a user-friendly, interactive AI tool that processes human-readable semantic labels about the desired expression manipulations in specific parts of the input video and synthesizes photorealistic manipulated videos. We achieve that by mapping the emotion labels to points on the Valence-Arousal space (where Valence quantifies how positive or negative is an emotion and Arousal quantifies the power of the emotion activation), which in turn are mapped to disentangled 3D facial expressions through an especially-designed and trained expression decoder network. The paper presents detailed qualitative and quantitative experiments, which demonstrate the effectiveness of our system and the promising results it achieves.
\end{abstract}

\section{Introduction}
Manipulation and synthesis of photorealistic facial videos is a significant challenge in computer vision and graphics. It has plenty of applications in visual effects, movie post-production for the film industry, video games, entertainment apps, visual dubbing, personalized 3D avatars, virtual reality, telepresence and many more fields. Concerning photographs, commercial  and research software allows editing colors and tone of photographs \cite{1.ashbrook2006adobe,2.ImageStyleTransfer} and even editing visual style \cite{3.luan2017deep}. Moreover, for manipulating facial videos, the traditional methods involve the use of physical markers on the faces and expensive setups of lights and cameras at different angles in controlled conditions, along with the use of complex CGI and VFX tools \cite{4.abouaf2000creating}. More robust, affordable, and data-driven approaches to learn from the facial features of the subject would allow easy and fine-grained control over manipulations and would be applicable even in videos captured “in-the-wild".

\begin{figure}[t]
\centering
\includegraphics[trim=0 52 0 50, clip, width=.9\textwidth]{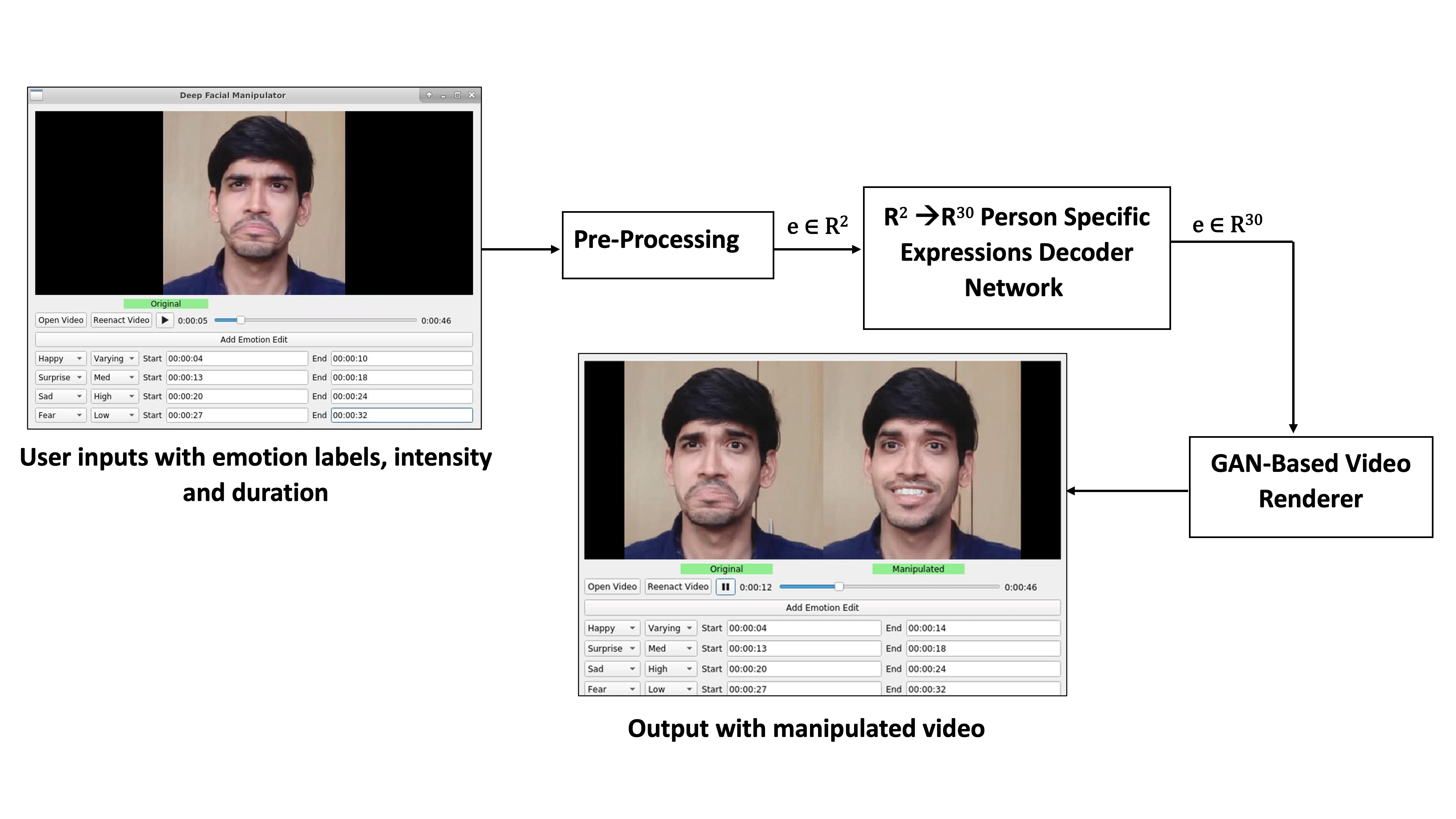}
\caption{Main steps of the proposed pipeline for semantic manipulation of facial videos.}
\label{fig:1}
\end{figure}

The vast majority of existing works perform manipulations of facial features in images rather than videos and typically require manual input by an experienced user. Some approaches are image-based and lack parametric control \cite{5.averbuch2017bringing,6.geng2018warp,7.zakharov2019few}, while other approaches like \cite{8.tewari2020stylerig} perform a parametric control by mapping parameters of a 3D Morphable Model (3DMM) to a generative model like the so-called StyleGAN \cite{9.karras2019style}. However, this disentangled control is on the images generated by Generative Adversarial Networks (GANs) and not on the original images. Tewari et al. \cite{10.tewari2020pie} overcome this limitation by developing an embedding for real portrait images in the latent space of the StyleGAN and support photorealistic editing on real images. The so-called ICface \cite{11.tripathy2020icface} makes use of Action Units (AUs) to manipulate expressions in images, producing good results, which however are not as visually plausible as the results produced by methods based on 3D models. Moreover, the manual control using AUs can be cumbersome due to too many parameters and the requirement of expert knowledge. Kollias et al.~\cite{kollias2020deep} synthesize visual affect from neutral facial images, based on input in the form of either basic emotion labels or Valence-Arousal (VA) pairs, supporting the generation of image sequences. Wang et al. \cite{44.wang2021facevid2vid} synthesize talking-head videos from a single source image, based on a novel keypoint representation and decomposition scheme. Even though this kind of methods output photorealistic videos and produce promising results, they solve a different problem than this paper, since they process input coming from just a single image and do not attempt to perform multiple semantic manipulations on a whole video of a subject. Gafni et al.~\cite{45.gafni2021nerface} use neural radiance fields for monocular 4D facial avatar reconstruction, which synthesize highly-realistic images and videos, having the ability to control several scene parameters. However, their method focuses on facial reenactment tasks and when used to manipulate a subject's facial video, the control over the facial expressions is rather limited and relies again on altering the parameters of a 3DMM.

This paper proposes a user-driven and intuitive method that performs photorealistic manipulation of facial expressions in videos. We introduce an interactive AI tool for such manipulations that supports user-friendly semantic control over expressions that is understandable to a wide variety of non-experts. Our AI tool allows the user to specify the desired emotion label, the intensity of expression, and duration of manipulation from the video timeline and performs edits coherent to the entire face area. We propose a novel robust mapping of basic emotions to expression parameters of an expressive 3D face model, passing through an effective intermediate representation in the Valence-Arousal (VA) space. 
In addition, we build upon state-of-the-art neural rendering methods for face and head reenactment \cite{12.koujan2020head2head,13.doukas2021head2head++}, achieving photorealistic videos with manipulated expressions.

\section{Related Work}

\subsubsection{Facial expression representation and estimation.}
Thanks to the revolution that Deep Learning and Convolutional Neural Networks (CNNs) have brought in the field, recent years have witnessed 
impressive advancements in video-based facial expression analysis \cite{19.Martnez2019AutomaticAO}. For example, Barros et al. \cite{20.barros2017emotion} use a deep learning model to learn the location of emotional expressions in a cluttered scene. Otberdout et al. \cite{21.otberdout2018deep} use covariance matrices to encode deep convolutional neural network (DCNN) features for facial expression recognition and show that covariance descriptors computed on DCNN features are more efficient than the standard classification with fully connected layers and SoftMax. Koujan et al. \cite{22.koujan2020real} estimate 3D-based representation of facial expressions invariant to other image parameters such as shape and appearance variations due to identity, pose, occlusions, illumination variations, etc. They utilize a network that learns to regress expression parameters from 3DMMs and produces 28D expression parameters exhibiting a wide range of invariance properties. Moreover, Toisoul et al. \cite{23.toisoul2021estimation} jointly estimate the expressions in continuous and categorical emotions.

\noindent \textbf{3D Morphable Models (3DMMs)} play a vital role in 3D facial modelling and reconstruction. They are parametric models and capable of generating the 3D representation of human faces. Following the seminal work of Blanz and Vetter \cite{24.blanz99}, the research in this field has been very active until today and during the last decade 3DMMs have been successfully incorporated in deep learning frameworks \cite{25.egger20203d}. Some recent advancements in this field produced more powerful and bigger statistical models of faces with rich demographics and variability in facial deformations \cite{26.huber2016multiresolution,27.dai20173d,28.FLAME:2017,29.booth2018large}.

\noindent \textbf{Facial reenactment} transfers facial expressions from a source to a target subject to conditioning the generative process on the source's underlying video. Traditional methods of facial reenactment use either  2D wrapping techniques \cite{31.liu2001expressive,32.garrido2014automatic} or a 3D face model \cite{33.thies2016face2face} manipulating only the face interior. 
More recent works are typically based on GANs, which are deep learning frameworks that have produced impressive results in image and video synthesis \cite{30.goodfellow2014generative}. Neural Textures \cite{34.thies2019deferred} perform real-time animation of facial expression of a target video by using an image interpolation technique to modify deformations within the internal facial region of the target and redirecting manipulated face back to the original target frame. There is no control over the head pose and eye gaze in face reenactment systems, but they are very useful in applications like video dubbing \cite{35.garrido2015vdub}. Kim et al.~\cite{36.NeuralStyle} further improve the realism of video dubbing by preserving the style of a target actor. They perform monocular reconstruction of source and target to use expressions of source, preserving identity, pose, illumination, and eyes of the target.

Many facial reenactment techniques exploit 3DMMs, since they offer a reliable way to separate expressions and identity from each other. For example, Deep Video Portraits \cite{37.kim2018deep} use a GAN-based framework to translate these 3D face reconstructions to realistic frames, being the first to transfer the expressions along with full 3D head position, head rotation, eye gaze and blinking. However, they require long training times, and the results have an unnatural look due to the inner mouth region and teeth. Head2Head \cite{12.koujan2020head2head} overcomes these limitations by using a dedicated multiscale dynamics discriminator to ensure temporal coherence, and a dedicated mouth discriminator to improve the quality of mouth area and teeth. This work was extended to Head2Head++\cite{13.doukas2021head2head++} by improving upon the 3D reconstruction stage and faster way to detect eye gaze based on 68 landmarks, achieving nearly real-time performance. Papantoniou et al.~\cite{papantoniou2022neural} build upon Head2Head++ and introduce a method for the photorealistic manipulation of the emotional state of actors in videos. They introduce a deep domain translation framework to control the manipulation through either basic emotion labels or expressive style coming from a different person's facial video. However, they do not exploit more fine-grained and interpretable representations of facial expressions, such as the VA representation adopted in our work. 

\begin{figure}[t]
\centering
\includegraphics[height=4cm]{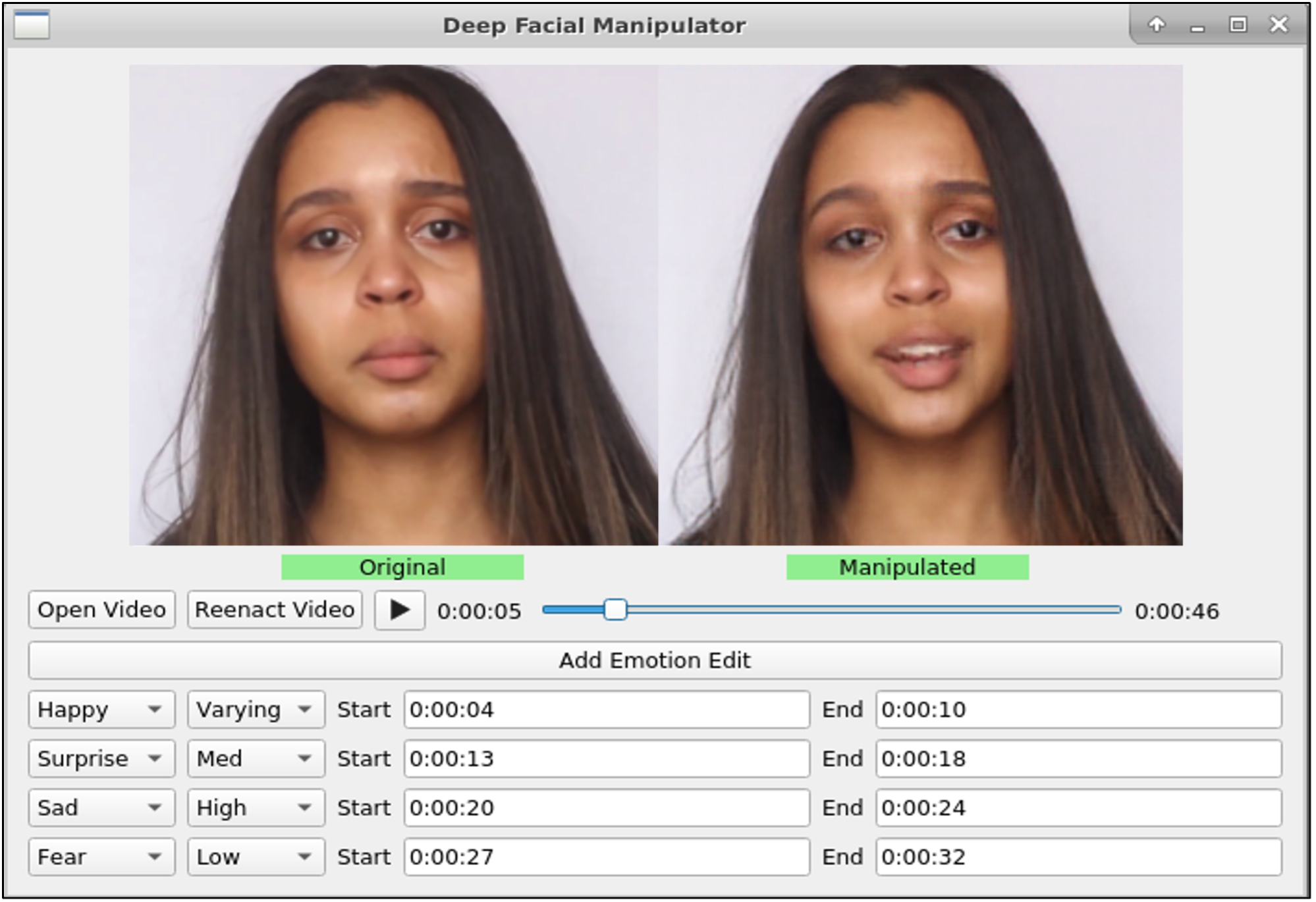}
\caption{GUI of our AI tool that processes semantic labels about the desired expression manipulations in specific parts of the input video and synthesizes photorealistic manipulated videos. The user can play the input and the manipulated video side by side to observe the effect of the manipulations. Please also refer to the supplementary video for a demonstration of the GUI \cite{suppmat}.}
\label{fig:2}
\end{figure}

\section{Proposed method}
We introduce a system that estimates and manipulates disentangled components of a 3D face from facial videos followed by self-reenactment, while maintaining a high level of realism. 
As depicted in Fig.~\ref{fig:1}, for every specific subject whose facial video we wish to manipulate, the main steps of our pipeline are as follows (more details are given in the following subsections):
\\

\noindent 
\textbf{1)} We collect several videos with a varied range of expressions for the specific subject. 
\textbf{2)} We process the collected videos and estimate for every frame of every video a data pair of VA values and Expression coefficients of a 3D face model by applying EmoNet \cite{23.toisoul2021estimation} and DenseFaceReg \cite{13.doukas2021head2head++} respectively. 
\textbf{3)} We use the data pairs generated in step 2 to train a person-specific Expression Decoder Network to learn a mapping from 2D VA values to 30D expression coefficients. 
\textbf{4)} For the video of the subject that we wish to manipulate, we train a state-of-the-art head reenactment method \cite{13.doukas2021head2head++} to produce a neural rendering network that produces photorealistic videos of the subject. 
\textbf{5)} To support the determination of semantic labels by the user, we assume input in the form of basic expression labels alongside their intensity, which we map to the VA space and from there to 3D expression coefficients, through the trained Expression Decoder Network (step 3). 
\textbf{6)} We input the manipulated 3D expression coefficients to the trained neural renderer (step 4), which outputs the manipulated video.


\subsection{AI tool}

As a Graphical User Interface (GUI) for the manipulations by the user, we introduce an interactive AI tool that we call “Deep Facial Manipulator” (DFM), Fig.~\ref{fig:2}. A user can open the video to be edited in the tool and can interactively add the edits by selecting a semantic label for the emotion as well as its intensity with start and end duration from the timeline. The “Reenact Video” button from the tool runs the reenactment process and the edited video is played in the tool once the process finishes. The user can then add more edits or modify the existing ones and repeat the process. Once the user is happy with the manipulations, they can keep the saved manipulated video. Please refer to the supplementary video for a demonstration of the usage of our AI tool \cite{suppmat}. The main steps that implement the AI tool’s functionality are described in more detail in the following sections.

\begin{figure}[t]
\centering
\includegraphics[height=2.7cm]{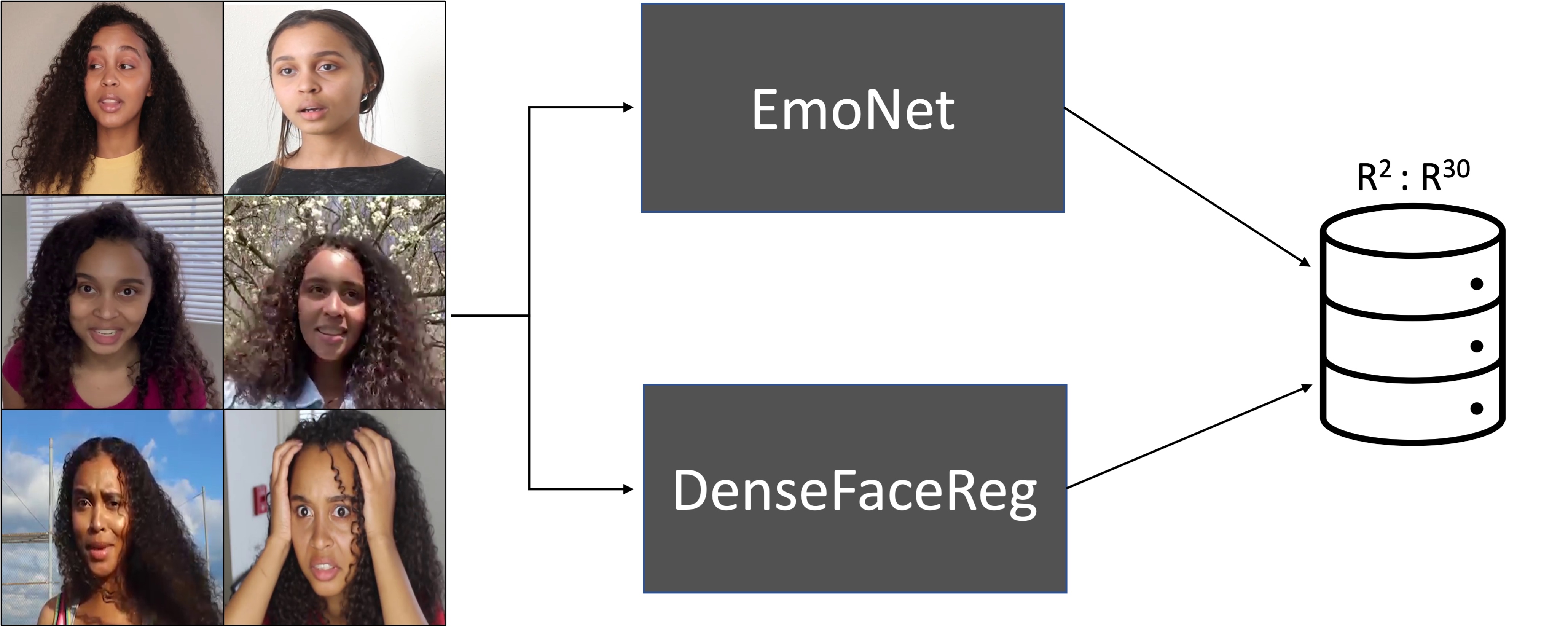}
\caption{Construction of a Person-Specific Dataset of Facial
Expressions. We process a collection of videos with varied
expressions and construct a large set of data pairs, where each pair consists 
of a 2D VA vector and a 30D expression
coefficients vector.}
\label{fig:3}
\end{figure}

\subsection{Person-specific dataset of facial expressions}\label{sec:ExpDataset}
\begin{figure}[t]
\centering
\begin{tabular}{cc}
\includegraphics[trim=0 40 0 70, clip, width=.39\textwidth]{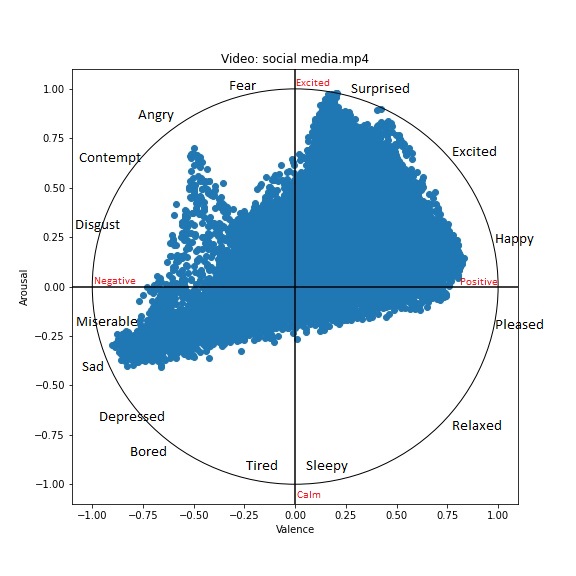} 
&
\includegraphics[trim=0 0 0 0, clip, width=.36\textwidth]{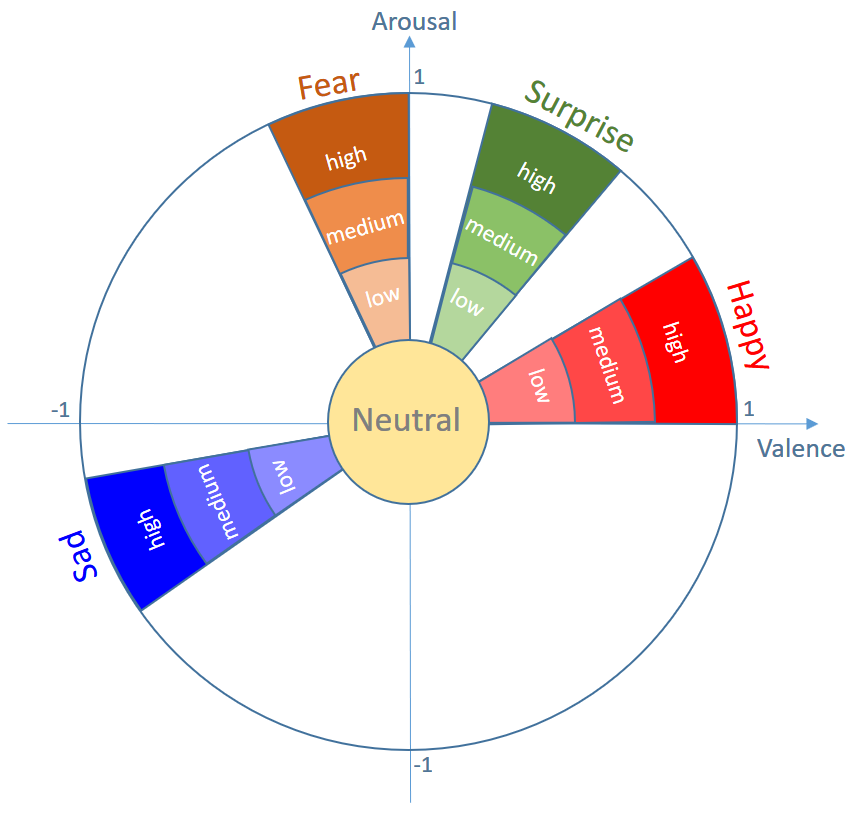}
\\
{\footnotesize	(a)}
&
{\footnotesize	(b)}
\end{tabular}
\caption{Adopted representation of facial expressions on the Valence-Arousal space. 
\textbf{(a)} Exemplar scatter plot of the Valence-Arousal points from one of our dataset videos. We observe that the points occupy a significant part of the VA space.
\textbf{(b)} Our pre-defined mapping from expression label and intensity to a specific region of the VA space, which we use to convert the user input in our AI tool. Please note that "Neutral" is the only label that does not have different strength levels.}
\label{fig:fig4}
\end{figure}

As already mentioned, for every subject whose video we want to manipulate, we collect a set of additional facial videos of the same subject with a varied range of expressions. This video collection helps our Expression Decoder Network learn the way that the specific subject is expressing their emotions. It is worth mentioning that the videos of this collection do not need to be recorded under the same conditions and scene as the video we wish to manipulate since we use and robustly estimate a disentangled representation of 3D facial expressions that is invariant to the scene conditions. For example, in the case that the subject is a celebrity actor, we can collect short clips from diverse interviews and different movie scenes with the specific actor. In more detail, we process every frame of every video of the person-specific collection to estimate the following (see Fig.~\ref{fig:3}):

\noindent\textbf{Valence-arousal values:} First, we
perform face detection and facial landmarking in every
frame by applying the state-of-the-art method of \cite{38.bulat2017far}. Then, we resize the cropped face
image to a resolution of 256x256 pixels and feed it to the
state-of-the-art EmoNet method \cite{23.toisoul2021estimation} to estimate Valence-
Arousal values. We have chosen Valence-Arousal values
as an intermediate representation of facial expression
because these provide a continuous representation that
matches more accurately the variability of real human
expressions, as compared to discrete emotion labels (like
Happy, Sad, Angry, etc.), while at the same time they can
be easily associated to expression labels specified by a nonexpert
user of our AI tool. Fig.~\ref{fig:fig4}(a) shows an example of
the distribution of VA values for one of the videos of the
collection of a specific subject.

\noindent\textbf{3D expression coefficients:} For every
frame, we perform 3D face reconstruction by running the
so-called DenseFaceReg network from  \cite{13.doukas2021head2head++}. This
is a robust and efficient CNN-based approach that
estimates a dense 3D facial mesh for every frame,
consisting of about 5K vertices. Following \cite{13.doukas2021head2head++}, we fit to
this mesh the combined identity and expression 3DMM
used in \cite{12.koujan2020head2head,13.doukas2021head2head++} and keep the 30-dimensional expression
coefficients vector that describes the 3D facial deformation due to facial expressions in a disentangled representation.
By performing the aforementioned estimations in the
person-specific video collection, we construct a dataset of
data pairs consisting of a 2D Valence-Arousal vector and a
30D expression coefficients vector.

\subsection{Expression decoder network}
\begin{figure}
\centering
\includegraphics[width=10cm]{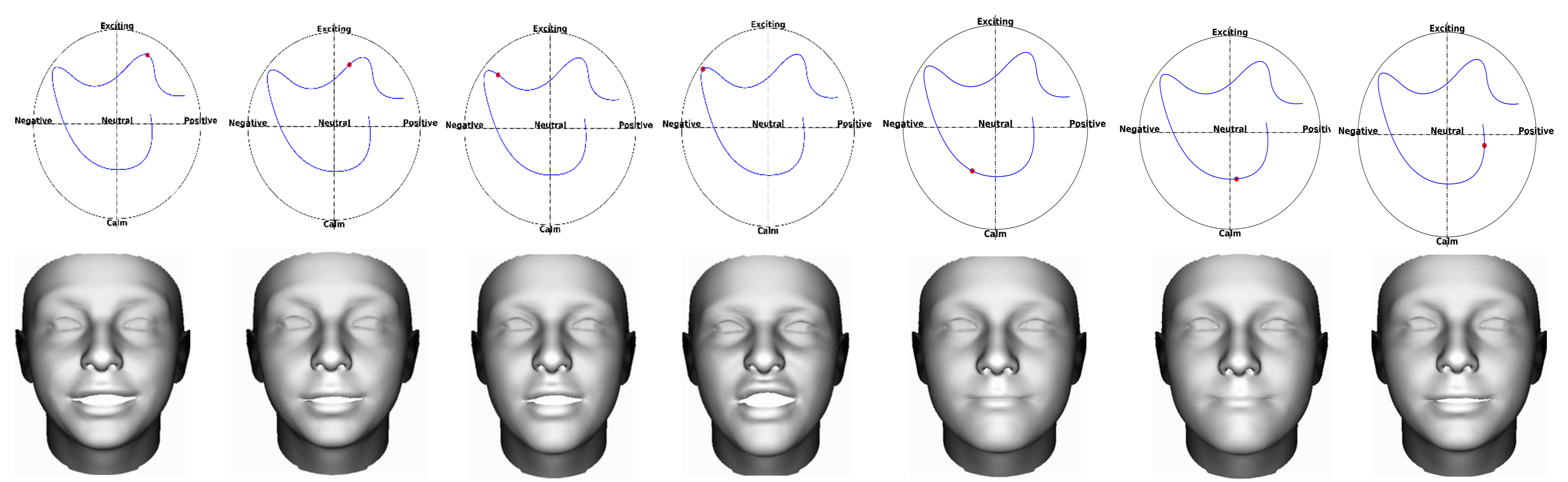}
\caption{Visualization of exemplar inference results of our Expression Decoder Network. $1^{st}$ row: sequence of input VA values, that form a 2D orbit, with the red dot signifying the current location. $2^{nd}$ row: corresponding output expression coefficients, visualized as a 3D mean face having the expression that is specified by these coefficients.}
\label{fig:fig5}
\end{figure}

Our Expression Decoder Network is a person-specific
mapping from the Valence-Arousal space to the expression
coefficients of the 3D face model. For every subject, we
train it in a supervised manner using the person-specific
dataset generated in Section \ref{sec:ExpDataset}. In more detail, we create a
multilayer perceptron network that takes a 2D VA vector and regresses a 30D expression coefficient
vector at the output. The loss function minimized during
training is the Root Mean Square Error (RMSE) loss
between the estimated and ground-truth expression
coefficients. The network consists of 6 fully-connected
layers with 4096, 2048, 1024, 512, 128 and 64 units per
layer respectively and with Rectified Linear Units (RELU)
to introduce non-linearities \cite{39.glorot2011deep}. The number of units and
layers are decided based on the empirical testing on a
smaller sample of data with different configurations and
was tuned by trying different learning rates and batch sizes.
A learning rate of $10^{-3}$ and batch size of 32 was selected,
based on the trials. To avoid overfitting and generalize
better, “dropout” is used as a regularization technique
where some network units are randomly dropped while
training, hence preventing units from co-adapting too
much \cite{40.srivastava2014dropout}. Furthermore, we used Adam optimizer \cite{41.kingma2015adam} for
updating gradients and trained the network for 1000
epochs. Figure~\ref{fig:fig5} demonstrates an example of results obtained by
our Expression Decoder Network (please also refer to the
supplementary video \cite{suppmat}). The input VA values
at different time instances form a 2D orbit in the VA space.
For visualization purposes, the output expression
coefficients provided by our Decoder Network are
visualized as 3D facial meshes by using them as parameters
of the adopted 3DMM \cite{12.koujan2020head2head,13.doukas2021head2head++} in combination with a mean
face identity. The visualized 3D meshes show that our
Decoder Network can generate a range of facial expressions
in a plausible manner.

\subsection{Synthesis of photorealistic manipulated
videos}

\begin{figure}[t]
\centering
\includegraphics[width=11cm]{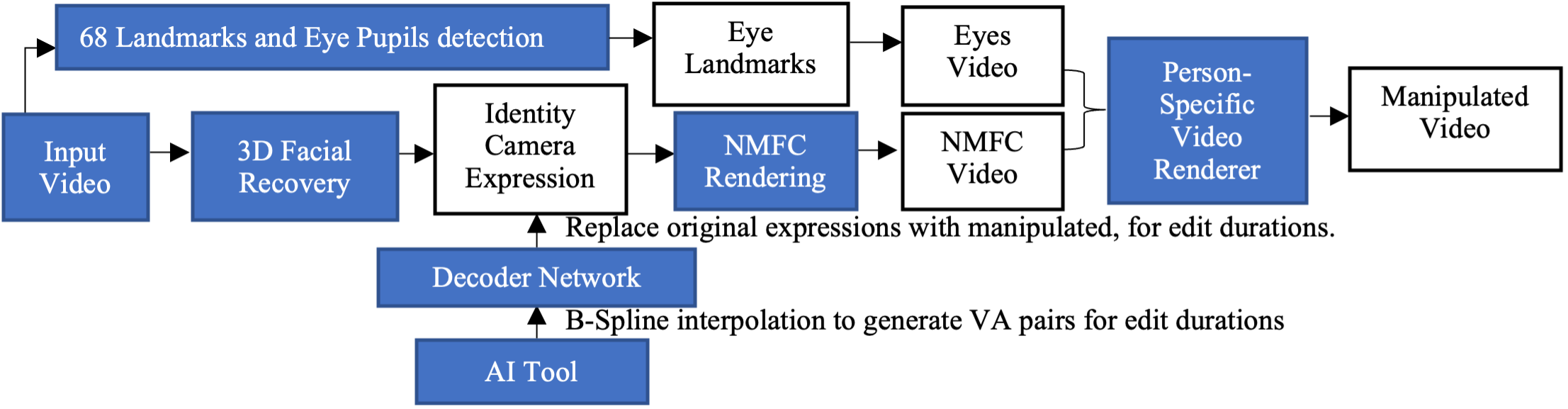}
\caption{Overview of our module for synthesis of photorealistic manipulated videos.}
\label{fig:fig6}
\end{figure}

An overview of our system’s module for synthesizing
photorealistic videos with manipulated expressions is
provided in Fig.~\ref{fig:fig6}. We are based on the
framework of GANs \cite{30.goodfellow2014generative}
and build upon the state-of-the-art method of Head2Head++
\cite{13.doukas2021head2head++}, which produces high-quality results in head
reenactment scenarios. We use the video to be manipulated
as training footage for Head2Head++ to train a neural renderer that synthesizes controllable sequences of fake
frames of the subject in the video to be manipulated. The
process involves the estimation of facial landmarks and eye
pupils, as well as disentangled components of the 3D face
(identity, expressions, pose), which helps us effectively
modify the expression component while keeping the other
components unaltered. These components are then
combined to create a compact image-based representation
called Normalized Mean Face Coordinate (NMFC) image,
which is further combined with the eye gaze video and used
as conditional input to the GAN-based video renderer to
synthesize the fake frames. Head2Head++ also takes care
of temporal coherence between frames and has a dedicated
mouth discriminator for a realistic reenactment of the mouth
region. 

Each facial expression manipulation is specified by the user in
the user-friendly format of a time interval of the edit,
emotion label, and intensity (low, medium, or
high). Adopting one of the two sets supported by EmoNet
\cite{23.toisoul2021estimation}, we consider the following 5 basic emotion labels:
neutral, happy, sad, surprise, fear. Inspired by \cite{23.toisoul2021estimation}, we use
a pre-defined mapping from emotion label and intensity to
a specific region of the VA space, see Fig.~\ref{fig:fig4}(b). To further 
improve the realism of the results, we randomly sample VA
values within this specific region and associate them with
equally spaced time instances within the time interval. We
use B-spline interpolation to connect these VA values and
create a sequence of VA values so that there is one VA pair
for every frame. These VA values are then fed to our Expression Decoder Network to generate the manipulated
expression parameters. To ensure a perceptually plausible
transition of expressions, in case that the desired expression
is different from neutral, we transit from neutral to the
desired expression and from the desired expression back to
neutral using 20 frames at the beginning and end of each
edit. Furthermore, to also ensure smooth dynamics of facial
expressions and avoid any noise and jittering artifacts, we
use approximating cubic splines for smoothing \cite{42.pollock1993smoothing}.

Having computed a vector of expression coefficients for
every frame to be manipulated, we combine it with the
identity and camera parameters from the original frame to
produce manipulated NMFC images, see Fig.~\ref{fig:fig6}. The
sequence of manipulated NMFC images is then utilized by
the trained neural renderer to synthesize
photorealistic manipulated frames with the desired
expressions.

\begin{figure}[t]
\centering
\includegraphics[width=7.6cm]{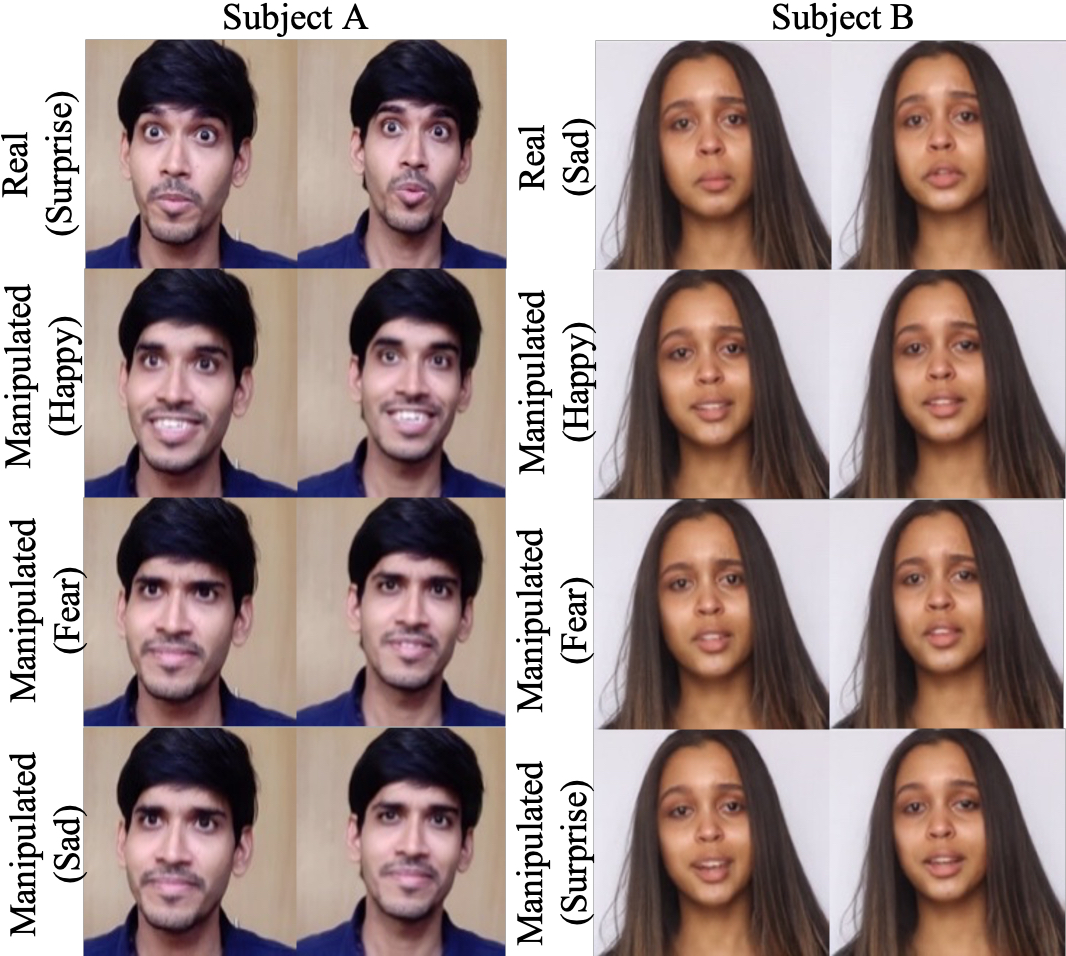}
\caption{Frames from exemplar manipulations using our deep semantic manipulation system. The output manipulated videos correspond to several different emotion labels.}
\label{fig:fig8}
\end{figure}

\section{Experimental results}
We conduct qualitative as well as quantitative
experiments to evaluate the proposed system. The
experiments presented in this section are conducted on
videos from 2 subjects: one male and one female (hereafter
referred to as subject A and subject B respectively - see
Fig.~\ref{fig:fig8}). All used videos have a frame rate of 30 fps. The
video collection for the person-specific datasets of facial
expressions (Sec.~\ref{sec:ExpDataset}) originated from videos shot with a
mobile camera for Subj. A (18 videos with an overall
duration of 1 hour 40 mins) and from publicly available
YouTube videos for Subj. B (20 videos with an overall
duration of 1 hour 55 mins). These video collections included variability in terms of facial expressions, to train our Expression Decoder Network. Furthermore, for each subject, the adopted neural renderer was trained on a single video (with 2 mins 17 sec of duration for subj. A and 1 min 23 sec of duration for subj. B), after following the pre-processing pipeline of \cite{13.doukas2021head2head++}, i.e.~face detection, image cropping and resizing to a resolution of 256x256 pixels. Please refer to the supplementary video for additional results and visualizations \cite{suppmat}.
\subsection{Qualitative evaluation}
We apply our method to generate several types of expression manipulations in videos, altering for example the emotion class from happy to sad, from neutral to happy, from sad to fear, and so on. 
Fig.~\ref{fig:fig8} illustrates examples of video frames with such manipulations, using different types of manipulated expressions. We observe that our method succeeds in drastically changing the depicted facial expressions in a highly photorealistic manner. In several cases, the altered facial expressions seem to correspond well to the target emotion labels (e.g., in the case of the label “happy”), but this is not always the case (e.g., in the label “sad”). This can be attributed to the different levels of availability and diversity of training data for different emotions.

\begin{figure}[t]
\centering
\includegraphics[trim=0 30 0 0, clip, width=.6\textwidth]{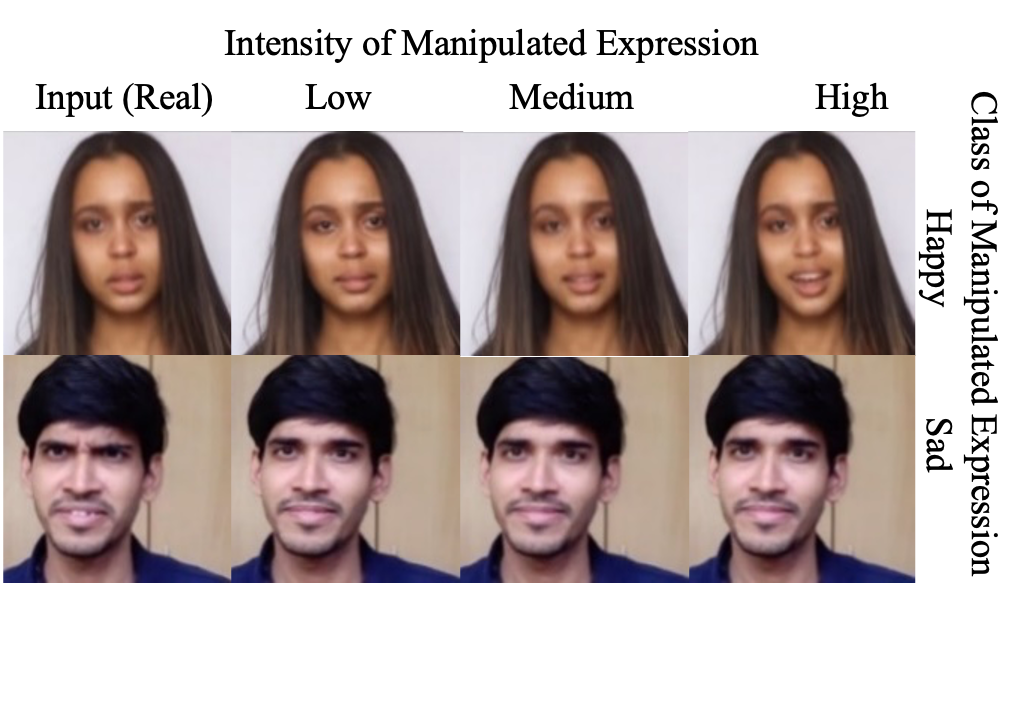}
\caption{Frames from exemplar manipulations using our system, where the target expressions and intensities are controlled. Please also see the supplementary video \cite{suppmat}.}
\label{fig:fig9}
\end{figure}

Fig.~\ref{fig:fig9} presents examples of controlling the intensity of the target expressions, which is also supported by the proposed method. It visualizes specific frames of the manipulated videos, with different manipulated expressions at different intensities (low, medium, and high). We observe that for the label “happy”, the effect of varying the controlled intensities is visually apparent, whereas for the label “sad” all three intensity levels yield very similar results. Furthermore, in the visualized example, the target label “happy” matches very well the perceived emotion by humans.

\subsection{Quantitative evaluation}
Following \cite{12.koujan2020head2head,13.doukas2021head2head++,37.kim2018deep} we quantitatively evaluate the synthesized videos through self-reenactment, since this is the only practical way to have ground truth at the pixel level. In more detail, the videos of subjects A and B that have been selected for training the person-specific video renderer undergo a 70\%-30\% train-test split (the first 70\% of the video duration is used for training and the rest for testing). The frames at the train set are used to train the adopted neural renderer as normal, whereas the frames at the test set are used to evaluate the self-reenactment and are considered as ground truth. The following pre-processing steps are applied to the test set: \textbf{i)} 3D face reconstruction to extract disentangled components (expression coefficients, identity, pose) for every frame, and \textbf{ii)} extraction of values for every frame using EmoNet \cite{23.toisoul2021estimation}. Subsequently, we test 2 types of self-reenactment: \textbf{A)} using “Ground Truth” expression coefficients, by which we mean that we feed our method with the expression coefficients extracted through the pre-processing step (i) described above. This type of self-reenactment corresponds to the one used in previous works \cite{12.koujan2020head2head,13.doukas2021head2head++,37.kim2018deep}. \textbf{B)} using expression coefficients synthesized by our Expression Decoder Network, after this is fed with valence-arousal values extracted through the pre-processing step (ii) described above. This is a new type of self-reenactment which we term as “emotion self-reenactment".  
Following \cite{13.doukas2021head2head++}, we use  the following metrics: \textbf{1)} Average Pixel Distance (APD), which is computed as the average L2-distance of RGB values across all spatial locations, frames and videos, between the ground truth and synthesized data. \textbf{2)} Face-APD, which is like APD with the only difference being that instead of all spatial locations, it considers only the pixels within a facial mask, which is computed using the NMFC image. \textbf{3)} Mouth-APD, which is the same as Face-APD but the mouth discriminator of Head2Head++ \cite{13.doukas2021head2head++}.

\begin{figure}[t]
\centering
\includegraphics[width=8.5cm]{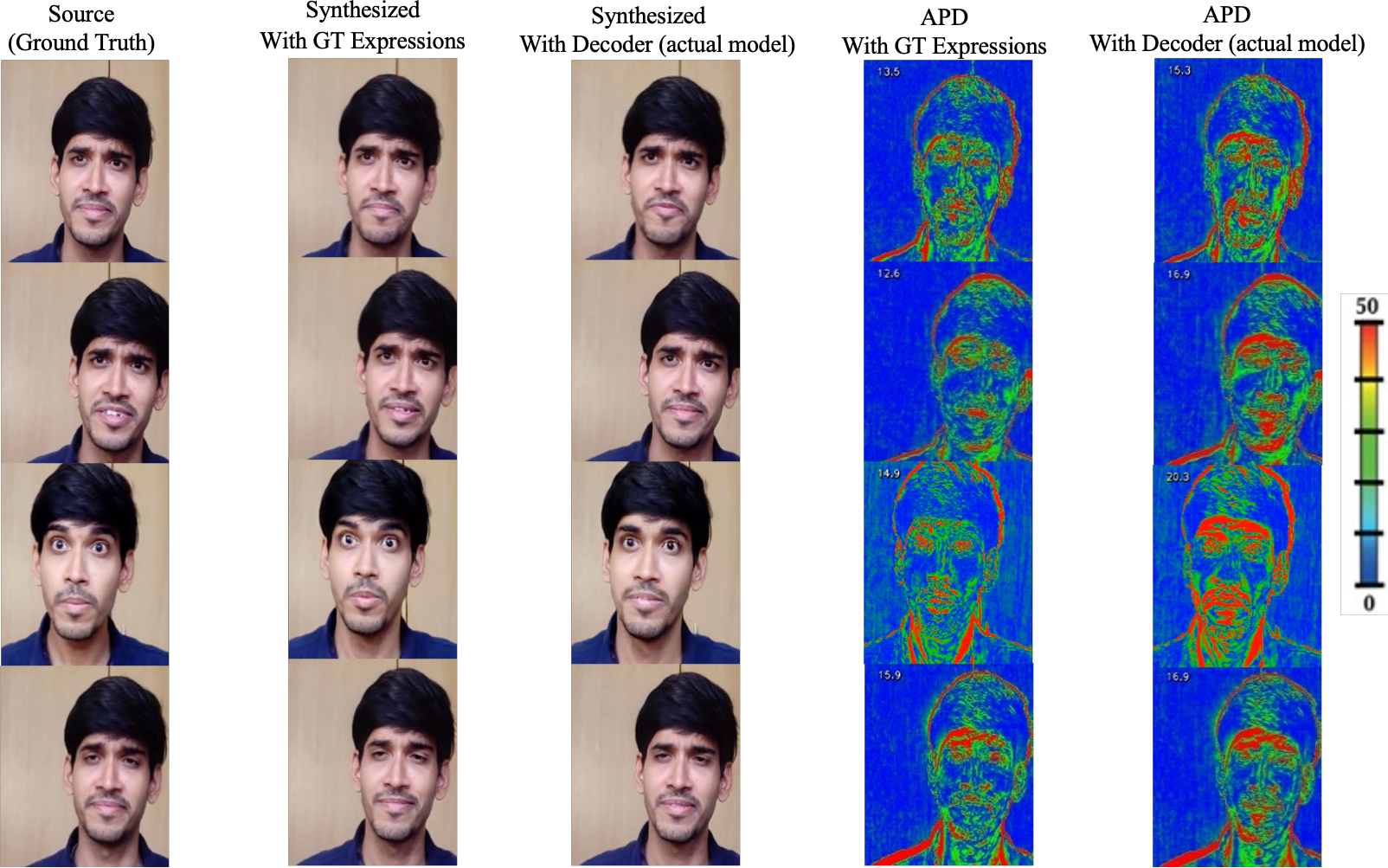}
\caption{Example frames from the self-reenactment evaluation. Our actual model is compared with the simplified case where the GT expressions are used. Despite solving a more difficult task, we achieve very similar quality, both visually and in terms of APD measures. We also observe that the synthesized frames are perceptually very similar to the source (ground truth). Please also see the supplementary video \cite{suppmat}.}
\label{fig:fig10}
\end{figure}

Fig.~\ref{fig:fig10} visualizes the synthesized images and APD metrics in the form of heatmaps for some example frames from the test set of subject A. We observe that our method manages to achieve accurate emotion self-reenactment. This is despite the fact that the synthesized expression of every frame comes from our Expression Decoder Network, using as sole information the valence-arousal values and without any knowledge about how these are instantiated in terms of facial geometry and deformations at the specific ground truth frame. We also see that the synthesized images as well as the values and spatial distribution of the APD errors is very close to the case where GT expressions are being used (“with GT expressions”), which solves the substantially less challenging problem of simple self-reenactment. Table~\ref{table:table1_APD} presents the overall quantitative evaluation in terms of APD metrics averaged over all pixels frames and both subjects. We observe that, unsurprisingly, as we move from APD to Face-APD and then to Mouth-APD, the error metrics increase, since we focus on more and more challenging regions of the face. However, all APD metrics of our method are relatively low and consistently very close to the case of simple self-reenactment (“with GT expressions”).

\begin{table}
\begin{center}
\caption{Quantitative evaluation of self-reenactment with GT expressions and with expression decoder network (our actual model). Average pixel distance (APD) values over all frames and over both subjects tested are reported. The considered range of pixel values is [0,255].}
\label{table:table1_APD}
\includegraphics[width=.95\textwidth]{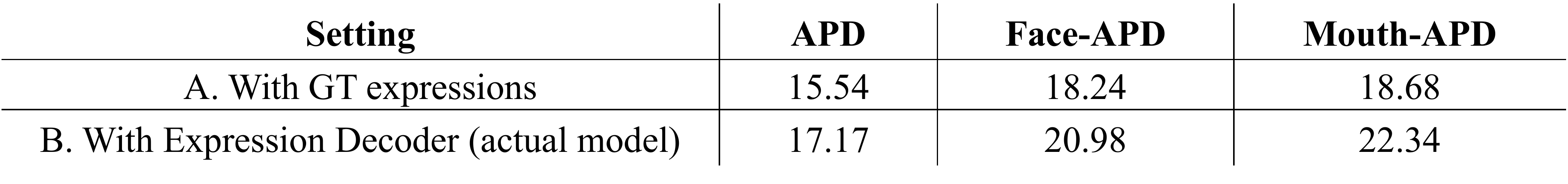}
\end{center}
\end{table}

\subsection{User studies}

\textbf{Realism.} We designed a comprehensive web-based user study to evaluate the realism of the manipulated videos, as perceived by human observers. Following \cite{12.koujan2020head2head,37.kim2018deep}, the questionnaire of the user study included both manipulated and real videos, with a randomly shuffled order and without revealing to the participants which videos were real. In addition, the participants were asked to watch each video once and rate its realism as per their first perception on a Likert-type scale of 1 to 5, where 1 means “absolutely fake” and 5 means “absolutely real”. Scores with values 4 or 5 are considered as corresponding to realistic videos. In total, 24short videos (with subjects A and B having an equal share of 12 videos each) were included in the questionnaire and shown to every participant of the user study. The manipulated videos were of two types: \textbf{A)} videos with single manipulation (SM), which corresponds to a single manipulated expression at the central part of the video (60\% of video duration) with a forward and backward transition from the real footage (real to fake and then back to real footage), and \textbf{B)} videos with double manipulation (DM), which included two manipulated expressions with similar forth and back transitions from real footage. In total, 21 participants took part in this user study.

Table~\ref{table:table2_Realism} presents the results of the user study on realism. We observe that for Subj. A, we achieve a realism score (percentage of “real”) that is almost the same with the score for real videos (77\% versus 79\%). This reflects the high realism of the results that our system is able to achieve. We also observe that for Subj. B, the performance of our system is relatively lower, which might be attributed to the fact that the training footage for this subject was relatively shorter. However, even for this subject, our system succeeds in synthesizing videos that are perceived as real more than 50\% of the times, which is still a promising result. In addition, we see that videos with single manipulation (SM) achieved better realism scores as compared with videos with double manipulation (DM). This is due to the fact that keeping a high level of realism becomes more challenging when multiple transitions to fake expressions are included in the manipulated video.

\begin{table}[t]
\begin{center}
\caption{Results of the user study about the realism of manipulated videos. 21 users participated in the study and the number of answers per score is reported. "Real" corresponds to the percentage of scores with values 4 or 5.}
\label{table:table2_Realism}
\includegraphics[trim=0 20 0 15, clip, width=.93\textwidth]{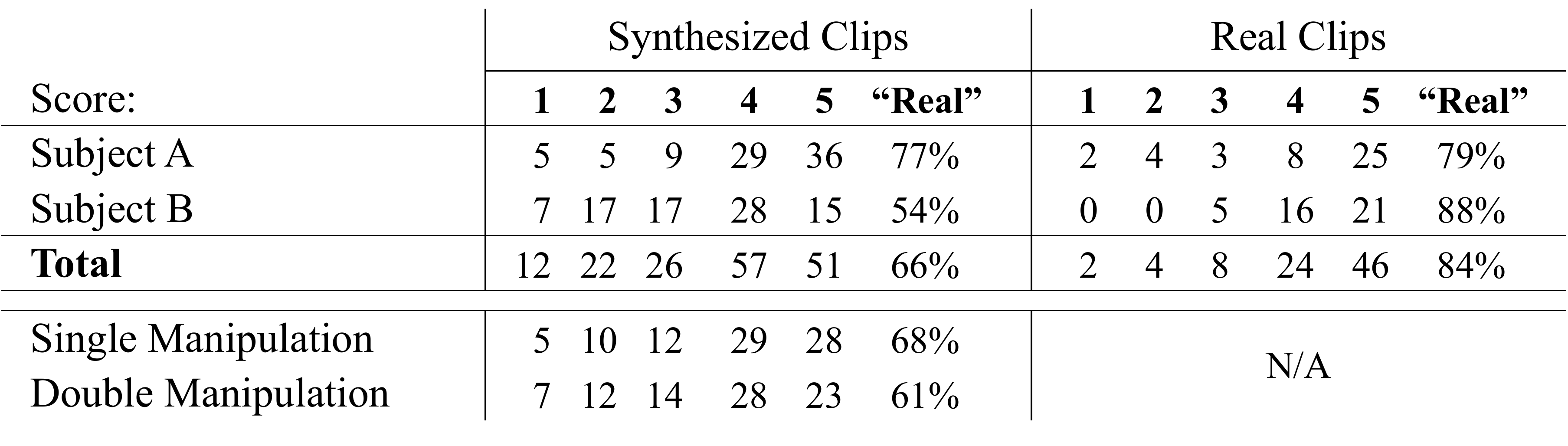}
\end{center}
\end{table}

\noindent\textbf{AI tool.} We designed another web-based user study, where we asked participants to try our AI tool “DFM”. The participants were provided with a small overview of the tool and were asked to operate it via a screen share session. The participants used the DFM tool to generate the manipulated videos and observe the results. Based on their user experience, they answered some questions regarding the tool and technology they used, based on a Likert-type scale from 1 to 5. The questions were related to the user experience in terms of usage, functionality, recommendation to a friend, and how much they were impressed by using such tool and technology. In total, 5 users took part in this study. 
The results of the user study for our AI tool are presented in Table~\ref{table:table3_tool}. We observe that all users responded very positively (scores 4 or 5) to the questions related to the ease of usage, the functionality, the chances of recommending the tool to a friend, and the excitement of the technology used. 
These results show the potential impact that our novel framework for photorealistic video manipulation can have. On the other hand, three users rated the speed and performance as 3 which is logical as users like faster results, and our implementation has not yet reached real-time speeds. In addition, the answers in the question regarding design and interaction show that some participants believe that there is a scope of improvement in the GUI design for a better user experience.

\begin{table}[t]
\begin{center}
\caption{
Results of the user study about our AI tool. The users rated different aspects of the tool on a Likert-type scale from 1 to 5. 5 users participated and the number of answers per question and score is reported.}
\label{table:table3_tool}
\includegraphics[width=.98\textwidth]{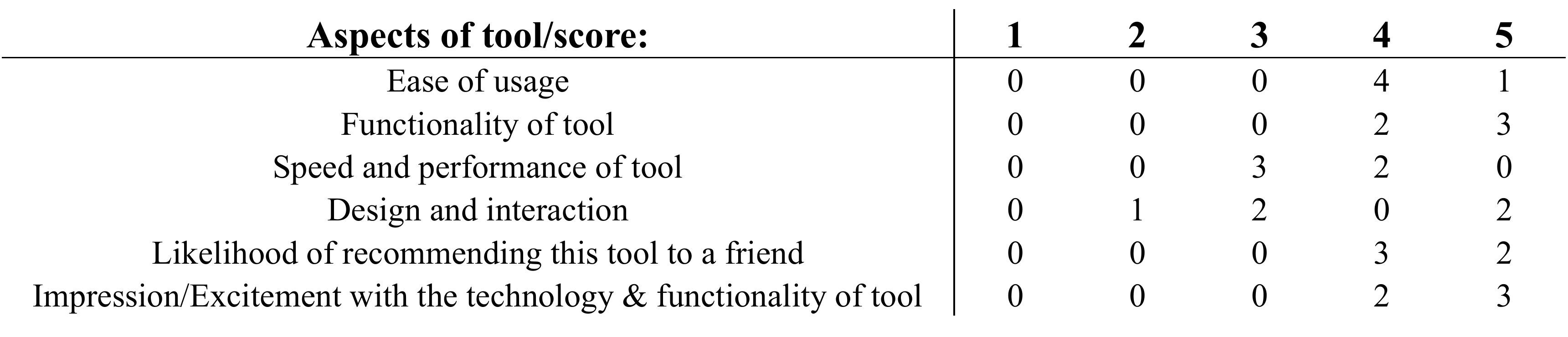}
\end{center}
\end{table}

\section{Conclusion and future work}
We proposed a novel approach to photorealistic manipulation of facial expressions in videos. Our system gives the ability to the user to specify semantically meaningful manipulations in certain parts of the input video in terms of emotion labels and intensities. We achieve this by mapping the emotion labels to valence-arousal (VA) values, which in turn are mapped to disentangled 3D facial expressions, through our novel Expression Decoder Network. In addition, we build upon the latest advances in neural rendering for photorealistic head reenactment. We also developed an easy-to-use interactive AI tool that integrates our system and gives the ability to a non-technical common user to access this functionality. The extensive set of experiments performed demonstrates various capabilities of our system and its components. Moreover, the qualitative and quantitative evaluations along with the results of the user studies provide evidence about the robustness and effectiveness of our method. 
For future work, we aim to increase the speed of the manipulations using the tool and eliminate the long training times of videos to be edited on the neural renderer. This seems achievable with recent advancements in works like Head2HeadFS \cite{43.christos2021head2headfs} that use few-shot learning. Furthermore, it might be interesting to develop a “universal” expression decoder network that will be able to achieve realistic results without needing to rely on such an extensive person-specific training. 
\\
\noindent \textbf{Note on social impact.} Apart from the various applications with positive impact on society and our daily lives, this kind of methods raise concerns, since they could be misused in a harmful manner, without the consent of the depicted individuals \cite{14.chesney2019deep}. 
We  believe that scientists working in these fields need to be aware of and seriously consider these risks and ethical issues. 
Possible countermeasures include contributing in raising public awareness about the capabilities of current technology and developing systems that detect deepfakes   \cite{15.de2020deepfake,16.amerini2020exploiting,17.korshunov2019vulnerability,18.Rssler2019FaceForensicsLT}. 
\\
\textbf{Acknowledgments.} A.Roussos was supported by HFRI under the `$1^{st}$ Call for HFRI Research Projects to support Faculty members and Researchers and the procurement of high-cost research equipment' Project I.C.Humans, Number 91.

\clearpage
%
%
\bibliographystyle{splncs04}
\bibliography{egbib}
\end{document}